\documentclass[11pt]{article}

\usepackage{acl}

\usepackage{times}
\usepackage{latexsym}
\usepackage{booktabs}
\usepackage{amsmath} 
\usepackage[T1]{fontenc}

\usepackage[utf8]{inputenc}

\usepackage{microtype}

\usepackage{inconsolata}

\usepackage{enumitem}

\usepackage{graphicx}
\usepackage{booktabs}
\usepackage{mdframed}
\usepackage{enumitem}
\usepackage{xcolor}
\usepackage{float}
\usepackage{fancyhdr}

\fancypagestyle{firstpagefooter}{
  \fancyhf{}

  \fancyfoot[C]{\footnotesize To appear in the Proceedings of the 21st Workshop on Innovative Use of NLP for Building Educational Applications (BEA 2026), co-located with ACL 2026. Copyright 2026 by the author(s).}
}

\title{Retrieval-Augmented Tutoring for Algorithm Tracing and Problem-Solving in AI Education}

\author{
  \textbf{Mragisha Jain\textsuperscript{1}},
  \textbf{Tirth Bhatt\textsuperscript{1}},
  \textbf{Griffin Pitts\textsuperscript{1}},
  \textbf{Aum Pandya\textsuperscript{1}},
\\
  \textbf{Peter Brusilovsky\textsuperscript{2}},
  \textbf{Narges Norouzi\textsuperscript{3}},
  \textbf{Arto Hellas\textsuperscript{4}},
  \textbf{Juho Leinonen\textsuperscript{4}},
  \textbf{Bita Akram\textsuperscript{1*}}
\\
\\
  \textsuperscript{1}North Carolina State University, 
  \textsuperscript{2}University of Pittsburgh, \\
  \textsuperscript{3}University of California, Berkeley,
  \textsuperscript{4}Aalto University,
\\
  \small{
    \textbf{*Correspondence:} \href{mailto:bakram@ncsu.edu}{bakram@ncsu.edu}
  }
}

\begin{document}
\maketitle
\thispagestyle{firstpagefooter}

\begin{abstract}
Students learning algorithms often need support as they interpret traces, debug reasoning errors, and apply procedures across unfamiliar problem instances. In this paper, we present KITE (Knowledge-Informed Tutoring Engine), a Retrieval-Augmented Generation (RAG)-based intelligent tutoring system designed to serve as a classroom teaching assistant for algorithmic reasoning and problem-solving tasks. KITE uses an intent-aware Socratic response strategy to tailor support to different student needs, responding with targeted hints, guiding questions, and progressive scaffolding intended to strengthen students’ algorithmic problem-solving ability. To keep responses aligned with course content, KITE uses a multimodal RAG pipeline that retrieves relevant information from course materials. We evaluate KITE using three forms of assessment: RAGAs-based metrics for response grounding and quality, expert evaluation of pedagogical quality, and a simulated student pipeline in which a weaker language model interacts with KITE across two-turn dialogues and produces revised answers after receiving feedback. Results indicate that KITE produces contextually grounded and pedagogically appropriate responses. Further, using simulated students, KITE’s feedback helped the student models produce more accurate follow-up responses on procedural and tracing questions, suggesting that its scaffolding can support algorithmic problem-solving. This work contributes a tutoring architecture and an evaluation approach for assessing retrieval-grounded explanations and scaffolded problem-solving feedback.
\end{abstract}

\section{Introduction}

Large language models (LLMs) such as ChatGPT are now widely used by students for learning support, including explanation, feedback, and problem-solving \cite{pitts2025student,pitts2026drives}. Students often value these tools because they provide immediate access to assistance when instructors or teaching assistants are unavailable \cite{pitts2025student}. Although these tools make information more accessible, prior work raises concerns that students may accept AI-generated responses without sufficient evaluation, especially when those responses appear complete and confident~\cite{instantresponsechatgpt,pitts2025students,pitts2026trust}. In education, this can lead students to bypass the reasoning processes that assignments are designed to develop~\cite{pitts2025students,pitts2026trust}. These concerns highlight the need for LLM-based systems that provide timely, course-grounded information while delivering pedagogically appropriate support that helps students reason through learning tasks.

Retrieval-Augmented Generation (RAG) \cite{rag} offers a promising approach for building course-grounded tutoring systems by allowing LLM responses to draw on curated instructional materials. This grounding can reduce unsupported or course-inconsistent claims and help align explanations with the concepts, terminology, and conventions used in a course. However, strong retrieval does not by itself ensure effective tutoring. Even when a system retrieves relevant material, it may still provide responses that are too direct, insufficiently instructional, or mismatched to the student’s immediate learning need. Prior work on intelligent tutoring systems suggests that effective support depends on both the accuracy of the information provided and how assistance is delivered, including when to offer direct explanation, feedback, or more guided support~\cite{koedinger2007exploring}.

Socratic tutoring offers one way to address this challenge by guiding learners through targeted questions, prompts, and progressive hints instead of immediately providing full solutions. This approach is grounded in cognitive apprenticeship and guided facilitation \cite{collins1989cognitive, hmelo2006goals}, and has been used in dialogue-based tutoring systems such as AutoTutor \cite{GRAESSER199935}. However, integrating Socratic guidance into retrieval-grounded tutoring systems remains an open design problem: a course-specific tutor must stay faithful to instructional materials while also providing feedback that fits the type of problem the student is trying to solve.

In this paper, we present \textbf{KITE} (Knowledge-Informed Tutoring Engine), a RAG-based intelligent tutoring system that connects students to relevant course materials while using intent-aware tutoring strategies to support different forms of help-seeking. KITE uses a multi-stage multimodal retrieval pipeline to locate relevant instructional content and an intent-aware response strategy to determine how that content should be used in the response. For questions that require direct explanation, KITE provides responses aligned with retrieved course materials. For procedural, debugging, validation, and tracing questions, KITE provides targeted feedback, guiding questions, and progressive hints to support student reasoning. To evaluate KITE, we first assess its retrieval-grounded outputs for non-procedural questions using RAGAs-based metrics for grounding, relevance, and response quality. We then evaluate procedural and tracing questions through a simulated student pipeline in which a weaker language model revises its answers after receiving KITE’s feedback. Finally, human experts assess the resulting interactions to judge feedback quality and whether the revised answers show improvement. This work contributes (1) KITE, an intent-aware tutoring system that combines multimodal retrieval with pedagogical support, and (2) an evaluation of its retrieval-grounded responses and scaffolded feedback using automated metrics, simulated students, and expert evaluation. We assess two research questions: \textbf{RQ1:} How well does KITE produce grounded, course-relevant responses for non-procedural student questions? and \textbf{RQ2:} To what extent does KITE’s feedback support improved responses on procedural and tracing questions?

\section{Related Work}

RAG-based educational assistants have been used for a range of instructional purposes, including interactive learning support, content generation, and large-scale course deployment \cite{LI2025100417}. Across these systems, grounding LLM responses in course materials has generally improved factual accuracy compared to unaugmented models. For example, KAG \cite{11391105} reports Precision@5 of 0.85 and a 34\% reduction in student task completion time, while MoodleBot \cite{article} achieves 88\% accuracy on course-related queries. However, these systems primarily function as direct question-answering tools and do not adapt their responses to different forms of student help-seeking.

Although RAG can improve factual accuracy, deployment studies suggest that course-grounded assistants also need evaluation in instructional workflows. In one classroom deployment, students showed strong pre-exam engagement but declining adoption across cohorts, and 36.8\% reported frustration when responses extended beyond a constrained knowledge base \cite{articlemed}. Edison \cite{inproceedings}, a GPT-4-based RAG assistant deployed in a large data science course, showed that retrieving from course documents and historical Q\&A can support factual and relevant responses to live student questions. The study also demonstrates the value of TA-in-the-loop evaluation, using instructor edits and ratings to assess factuality, relevance, style, and efficiency. EduMod-LLM \cite{mittal2026edumod} extends this line of work by treating educational Q\&A as a modular pipeline, separating function calling, retrieval, and response generation so that system behavior can be evaluated more transparently. 

Dialogue-based tutoring provides another foundation for supporting student reasoning. AutoTutor \cite{GRAESSER199935} showed that progressive hints and collaborative answer refinement produced dialogues rated above the ``good'' threshold by domain experts, with semantic evaluation correlating at 0.49 with expert judgment. More recently, \cite{LI2026104770} reported significant gains in self-efficacy ($d = 0.57$) from a Socratic AI platform in healthcare education. LeanTutor \cite{patel2026leantutor} similarly emphasizes guided feedback by combining LLMs with a theorem prover to check student proofs, identify errors, and provide hints toward a correct proof without giving away the complete answer. These systems show the value of scaffolded feedback for learning tasks that require students to reason through a process, yet they do not incorporate retrieval grounding to keep responses aligned with course-specific materials. 

\begin{figure*}[t]
    \centering
    \includegraphics[width=0.74\linewidth]{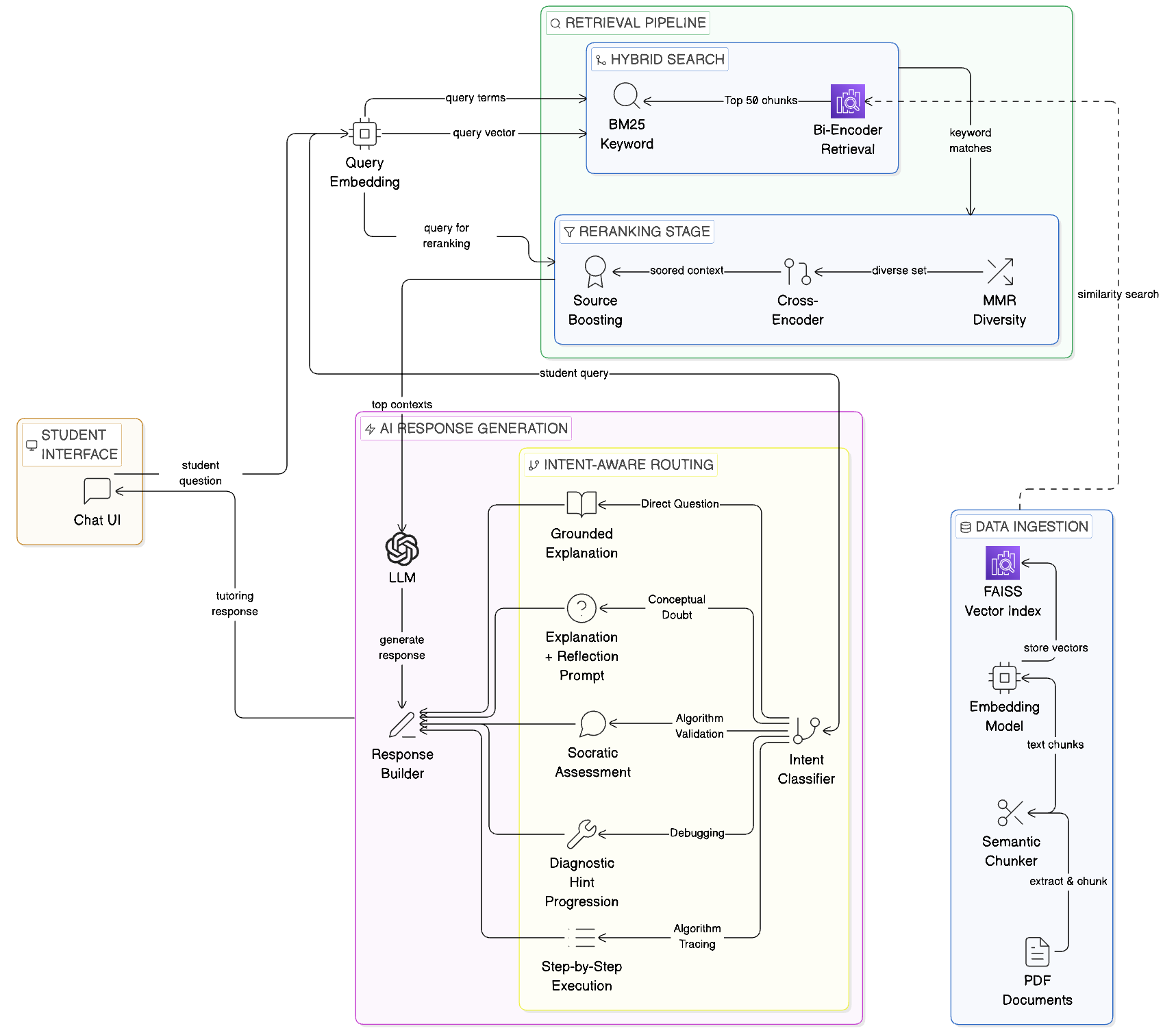}
    \caption{KITE architecture}
    \label{fig:kite_architecture}
\end{figure*}

Other systems explore how retrieval and response strategies can be adapted for learning contexts. LPITutor \cite{Liu2025LPITutorAL} supports adaptive difficulty modulation through RAG and prompt engineering. KG-RAG \cite{10975937} combines semantic retrieval with an expert-validated knowledge graph and reports a 35\% improvement in learning outcomes ($d = 0.86$) in a study of 76 students, though its reliance on manual expert validation limits scalability. AutoTA \cite{autoTA} provides a related approach to intent-aware educational assistance by classifying student queries and routing them to specialized response strategies. These systems show the value of adapting retrieval, domain structure, and response behavior to different learning needs. KITE builds on this direction by pairing multi-stage retrieval with intent-aware tutoring strategies for algorithmic reasoning tasks.

\section{System Design}

KITE is a retrieval-augmented tutoring system designed to support course-grounded dialogue for algorithmic reasoning and problem-solving tasks. As shown in Figure \ref{fig:kite_architecture}, the system includes five components: document preprocessing, embedding generation, multi-stage retrieval, intent-aware response generation, and session management.

\subsection{Phase 1: Document Ingestion and Preprocessing}

KITE begins by extracting text from course PDFs using \textit{PyMuPDF}. Extraction is performed page by page so that the original document structure remains traceable during retrieval. To reduce noise before indexing, the system applies a \textbf{frequency-based} cleaning procedure that removes repeated headers, footers, page numbers, and other formatting artifacts. Specifically, it examines the first and last two lines of each page, identifies repeated patterns that occur across pages, removes those patterns along with page numbers and special characters, and normalizes whitespace.

The cleaned text is then segmented into semantically coherent chunks for retrieval. We use \textbf{section-aware chunking} with a target size of 500 characters, or about 125 tokens, and a 100-character overlap. Headers are identified and retained to preserve local structure, while overlap carries forward the final two sentences of the preceding chunk.

\subsection{Phase 2: Embedding Generation}

Each chunk is encoded using OpenAI's \texttt{text-embedding-3-large} model, producing 3072-dimensional embeddings. These vectors are L2-normalized so cosine similarity reflects semantic direction and are stored in a FAISS index \cite{FAISS} for efficient local retrieval.

\subsection{Phase 3: Multi-Stage Retrieval Pipeline}

KITE uses a multi-stage pipeline designed to balance high recall and precision in retrieving course content. Retrieval begins with a dense bi-encoder search that returns the top 50 candidate chunks for a given student query. The query and document chunks are encoded independently, and similarity is computed using cosine similarity, allowing the system to capture semantically related content. 

The candidate set is then refined through hybrid retrieval. Dense similarity contributes 70\% of the retrieval score, while sparse BM25 keyword matching contributes 30\%. This combination captures both semantic similarity and exact lexical overlap, which is useful when students use course-specific terminology, notation, or algorithm names.

To reduce redundancy among retrieved passages, KITE applies Maximal Marginal Relevance (MMR) with $\lambda$ set to 0.7:
\[
MMR = \lambda \times \text{Relevance} + (1-\lambda) \times \text{Diversity}
\]

The retrieved candidates are reranked using a \texttt{cross-encoder/ms-marco-MiniLM-L-6-v2} reranking model implemented through Sentence Transformers, where the query and document are jointly encoded to produce more precise relevance scores. Finally, KITE applies source-based boosting so that chunks from official course materials receive higher priority. Chunks with reranking scores above 0.6 receive an additional boost of 0.3. The final context passed to the generator consists of the top eight retrieved chunks.

\subsection{Phase 4: Intent Classification and Pedagogical Response Generation}

KITE does not use a single response strategy for all student questions. Instead, it first classifies each query by pedagogical intent and then generates a response that matches the instructional purpose of the interaction. This allows the system to distinguish among questions, debugging requests, and other forms of help-seeking.

\subsubsection{Intent Classification}

Each incoming query is classified into one of five pedagogical intents using a keyword and pattern-matching classifier, as shown in Figure \ref{fig:kite_architecture}.

  \begin{itemize}
  \setlength{\itemsep}{0pt}
  \setlength{\parskip}{0pt}
  \setlength{\parsep}{0pt}
  \setlength{\topsep}{0pt}
  \item \textbf{Direct Question}: factual queries seeking definitions or
    explanations (e.g., ``What is A*?'').
  \item \textbf{Conceptual Questions}: deeper \textit{why} or \textit{how} questions
    probing understanding (e.g., ``Why does BFS guarantee shortest path?'').
  \item \textbf{Algorithm Validation}: queries where a student submits their own
    implementation or trace for assessment.
  \item \textbf{Debugging}: queries involving a specific error or incorrect output.
  \item \textbf{Algorithm Tracing}: requests to step through the execution of an
    algorithm on a concrete problem instance
    (e.g., ``Trace A* on this graph starting from node S'').
  \end{itemize}

The classified intent determines which response generation strategy is invoked. In addition to these five query intents, KITE includes a dedicated answer evaluation mode for cases in which a student submits a written answer for assessment. This mode bypasses intent classification and routes directly to the feedback generation pipeline.

\subsubsection{LLM Generation and Intent-Aware Response Strategy}

All response generation in KITE is handled by GPT-5. Outputs are grounded in a structured prompt that injects the top eight retrieved chunks into a \texttt{[CONTEXT]} block. The model is instructed to prioritize course materials and avoid introducing information that is unsupported by the retrieved context, helping keep responses aligned with the course.

For direct questions and conceptual doubts, KITE produces explanations grounded in the retrieved material. Responses are written in a tutor-like tone that emphasizes reasoning instead of brief answer delivery. For conceptual doubts, the response also includes a follow-up question intended to prompt reflection.

For algorithm validation tasks, KITE adopts a Socratic assessment strategy instead of directly identifying errors. Responses include a brief evaluation of the student’s approach, acknowledgement of correct components, and guiding questions that target specific issues. This design supports learning without explicitly revealing the final solution, encouraging students to continue working through the problem independently.

For debugging assistance, KITE generates diagnostic prompts that guide students toward identifying errors through self-examination. Each response follows a structured hint progression and includes a learning point that connects the observed bug to the underlying conceptual principle, reinforcing understanding beyond the immediate correction.


For algorithm tracing queries, KITE retrieves the relevant procedural steps and rules from course materials and applies them step by step to the student’s specific problem instance. Each step explicitly maintains and updates algorithmic state variables such as OPEN lists, CLOSED sets, and selected nodes, following the tie-breaking rules and constraints defined in the query. The response concludes with the final path and cost when applicable.



\subsection{Phase 5: Session Management}

KITE maintains session state across multi-turn interactions to preserve continuity within a conversation. For each session, the system stores the original query, detected intent, prior responses, and any hints provided. When a student submits a follow-up query, KITE uses this stored context to determine how the interaction should continue: related direct and conceptual questions are treated as follow-ups, while validation, debugging, and tracing requests remain within their intent-specific response strategy when they concern the same problem or algorithm. For these turns, KITE constructs a brief context summary from the prior interaction and appends it to the retrieval prompt to reduce repetition and support progressive guidance.




\section{Methodology}

To evaluate KITE, we use three forms of assessment. First, we examine non-procedural responses using RAGAs-based metrics for grounding, relevance, and answer quality against instructor-authored reference answers. We then use a simulated student pipeline to assess whether KITE’s feedback helps produce improved responses on procedural and tracing questions. Finally, experts evaluate the pedagogical quality of KITE’s feedback and the resulting answer revisions.

  \begin{figure*}[t]
      \centering
      \includegraphics[width=.95\textwidth]{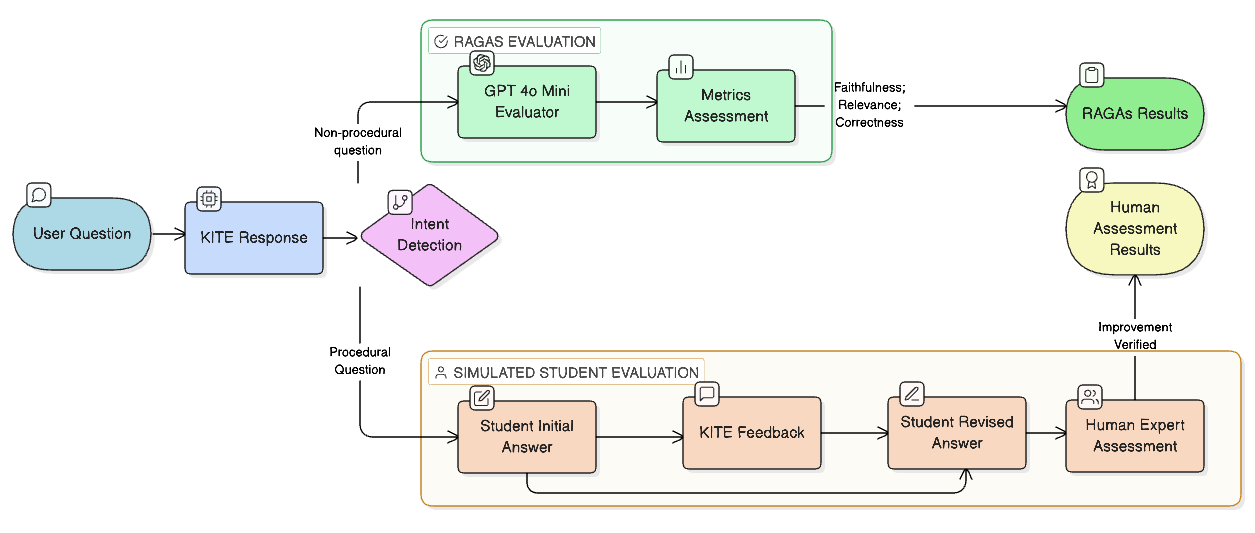}
      \caption{Evaluation pipeline}
      \label{fig:evaluation_pipeline}
  \end{figure*}

\subsection{Evaluation Dataset}

We constructed an evaluation dataset of 109 questions drawn from the lecture slides and textbook used in a university \textit{Introduction to AI} course, with each question paired with an instructor-verified reference answer. The dataset included 42 algorithmic questions, 51 procedural questions, and 16 direct-retrieval questions. We applied RAGAs to the 58 non-procedural questions, consisting of the algorithmic and direct-retrieval subsets, because these responses can be evaluated against reference answers for grounding, relevance, and answer quality. Questions requiring procedural reasoning or algorithm tracing were evaluated separately through the simulated student pipeline and expert review described in Section~\ref{sec:track2}.

\subsection{RAGAs Evaluation} \label{sec:track1}

We evaluate KITE’s non-procedural responses using the RAGAs framework~\cite{es-etal-2024-ragas,roychowdhury2024evaluation}, reporting six metrics:

\begin{itemize}
\setlength{\itemsep}{0pt}
\setlength{\topsep}{0pt}

\item \textbf{Faithfulness}: Measures whether statements in the generated response are supported by the retrieved context, computed as the proportion of answer claims judged to be grounded in the retrieved chunks.

\item \textbf{Answer Relevance}: Measures how well the generated response addresses the original question, computed from the cosine similarity between the user’s question and questions generated from the response.

\item \textbf{Context Relevance}: Measures how much of the retrieved context is relevant to answering the question, computed as the proportion of retrieved statements judged to be useful.

\item \textbf{Answer Similarity}: Measures semantic similarity between the generated response and the instructor-authored reference answer using sentence embeddings.

\item \textbf{Factual Correctness}: Measures factual agreement between the generated response and the reference answer using an F1 score over claims classified as true positives, false positives, and false negatives.

\item \textbf{Answer Correctness}: Measures overall correctness of the generated response relative to the reference answer as a weighted combination of factual correctness (0.75) and answer similarity (0.25).

\end{itemize}
  
All metrics use \texttt{gpt-4o-mini} as the judge model and \texttt{text-embedding-3-small} for similarity metrics. Retrieval uses \texttt{top\_k=5} from an initial candidate pool of 50.

\subsection{Simulated Student Evaluation and Expert Evaluation}
\label{sec:track2}

For procedural and algorithm-tracing questions, KITE provides Socratic feedback and guidance, making standard automatic scoring less appropriate. To evaluate how well this feedback supports learning-oriented revision, we use a two-stage simulated student pipeline followed by expert review.

\paragraph{Simulated Student Pipeline.}

Building on prior work that uses simulated student-tutor interactions to evaluate pedagogical support \cite{simulatedstudent}, we use Meta-Llama-3.1-70B-Instruct as a proxy student in a structured interaction with KITE:

\begin{enumerate}
\setlength{\itemsep}{0pt}
\setlength{\topsep}{0pt}
\item \textbf{Round 1}: The student model answers each question without assistance, establishing an unaided baseline.
\item \textbf{KITE Feedback}: KITE evaluates the student’s answer and provides feedback intended to guide revision.
\item \textbf{Round 2}: The student model revises its answer using KITE’s feedback.
\end{enumerate}

\paragraph{Expert Evaluation.}

Three experts reviewed each interaction set, consisting of the Round 1 answer, KITE’s feedback, and the Round 2 answer. They judged whether the revised response demonstrated improved correctness and reasoning, and evaluated the quality of KITE’s feedback using a structured rubric adapted from prior work \cite{mavrikis2025automating}.

The rubric includes three dimensions. \textit{Mistake Remediation} assesses whether the tutor correctly identifies the student’s error and explicitly acknowledges it in the response. \textit{Scaffolding and Guidance} assesses whether the tutor provides appropriate support without revealing the answer and offers clear next-step direction. \textit{Coherence and Tone} assesses whether the dialogue reads naturally and maintains an encouraging and supportive tone. Each criterion is scored as Yes/No, with NA used when a criterion is not applicable.



\section{Results}

\subsection{RAGAs Evaluation}

Table~\ref{tab:ragas} reports the six RAGAs metrics evaluated on the 58 non-procedural questions, consisting of 42 algorithmic questions and 16 direct-retrieval questions.

\begin{table}[h]
\centering
\begin{tabular}{lcc}
\toprule
\textbf{Metric} & \textbf{Mean} & \textbf{Std.\ Dev.} \\
\midrule
Faithfulness          & 0.8486 & 0.2103 \\
Answer Relevance      & 0.7558 & 0.2032 \\
Context Relevance     & 0.9352 & 0.1905 \\
Answer Similarity     & 0.7586 & 0.0923 \\
Factual Correctness   & 0.4483 & 0.2477 \\
Answer Correctness    & 0.6363 & 0.1810 \\
\bottomrule
\end{tabular}
\caption{RAGAs evaluation summary ($n = 58$).}
\label{tab:ragas}
\end{table}

KITE performs strongly on retrieval and grounding metrics. Faithfulness (0.85) indicates that most answer statements are supported by the retrieved context, while context relevance (0.94) shows that the retrieved passages are highly pertinent to the question. Answer relevance (0.76) and answer similarity (0.76) further suggest that KITE’s responses remain on-topic and semantically aligned with instructor-authored reference answers.

Factual correctness (0.45) is lower than the other RAGAs measures. As discussed in Section~7, this metric is sensitive to claim-level overlap with a single reference answer and may understate the quality of responses that are accurate but phrased differently or provide additional valid detail. For this reason, answer similarity is used as the primary indicator of response quality in this setting. Its low variance ($\sigma = 0.09$) also suggests relatively consistent performance across the evaluated questions.

\subsection{Simulated Student and Expert Evaluation}

Table~\ref{tab:human} summarizes the expert rubric scores for 44 simulated student--KITE interaction triples. Inter-rater agreement between the two expert annotators was high, with Cohen's $\kappa = 0.88$ and a raw agreement rate of 98.15\%, indicating strong consistency in rubric judgments.
 
\begin{table}[h]
\centering
\resizebox{\columnwidth}{!}{%
\begin{tabular}{lccc}
\toprule
\textbf{Metric} & \textbf{\% Yes} & \textbf{\% No} & \textbf{\% N/A} \\
\midrule
Mistake Remediation (Identifying)   & 63.63 & 6.82 & 29.55 \\
Mistake Remediation (Acknowledging) & 63.63 & 6.82 & 29.55 \\
Scaffolding                         & 93.18 & 6.82 & ---   \\
Guidance                            & 93.18 & 6.82 & ---   \\
Coherence (Naturalness)             & 93.18 & 6.82 & ---   \\
Tone (Encouraging)                  & 93.18 & 6.82 & ---   \\
\bottomrule
\end{tabular}}
\caption{Expert evaluation rubric results ($n=44$).}
\label{tab:human}
\end{table}

KITE receives consistently high ratings for scaffolding, guidance, coherence, and tone, with 93.18\% Yes judgments on each dimension. These results indicate that its feedback is generally well-structured, actionable, and supportive throughout the interaction. Mistake remediation receives 63.63\% Yes judgments, but 29.55\% of cases are marked N/A because the simulated student’s initial response was already correct and no error identification was required. When remediation is applicable, the results indicate that KITE identifies and acknowledges student errors appropriately.

\paragraph{Answer Improvement.}

Table~\ref{tab:improvement} reports how students responses changed from Round~1 to Round~2 after receiving KITE’s feedback. The transition labels reflect expert judgments of whether responses were Incorrect, Partially Correct, or Correct with respect to the course materials. Among the 27 interactions that were not already correct, 24 showed improvement after KITE’s feedback (88.89\%).

The most common transition was from Partially Correct to Correct, occurring in 14 cases (31.82\%). This suggests that KITE is particularly effective at helping students resolve remaining reasoning gaps in responses that are already moving in the right direction. In six additional cases (13.63\%), the response remained Partially Correct but still improved in quality, indicating that KITE’s feedback can support meaningful revision even when the student model does not reach a correct answer.

\begin{table}[h]
\centering
\resizebox{\columnwidth}{!}{%
\begin{tabular}{lcc}
\toprule
\textbf{Transition} & \textbf{Count} & \textbf{\%} \\
\midrule
Incorrect $\rightarrow$ Correct                                    &  1 &  2.27 \\
Incorrect $\rightarrow$ Partially Correct                          &  3 &  6.82 \\
Already Correct                                                     & 17 & 38.64 \\
Partially Correct $\rightarrow$ Correct                            & 14 & 31.82 \\
Partially Correct $\rightarrow$ Partially Correct with Improvement &  6 & 13.63 \\
N/A                                                                 &  3 &  6.82 \\
\bottomrule
\end{tabular}}
\caption{Answer improvement breakdown ($n = 44$).}
\label{tab:improvement}
\end{table}

\section{Discussion}
This study examined whether a course-grounded, intent-aware tutoring system could provide reliable retrieval-based support and pedagogically useful feedback for problem-solving tasks. The results are encouraging with regard to both aims. For RQ1, KITE’s faithfulness (0.85) and context relevance (0.94) indicate that its responses are closely grounded in retrieved course material, while answer similarity (0.76) shows consistent alignment with instructor-authored reference answers. For RQ2, among the 27 simulated-student interactions with KITE, in which the initial student response was not already correct, experts judged 24 revised answers (88.89\%) as improved after receiving KITE's feedback. This suggests that KITE’s feedback provided guidance the student model could use to correct or strengthen its reasoning in a follow-up response. Experts rated 93.18\% of KITE's feedback positively for scaffolding and guidance, further indicating that the feedback was instructionally purposeful and well-structured.

\textbf{Retrieval and response quality.} The RAGAs evaluation indicated that KITE performed well on measures tied to retrieval and grounding, with context relevance of 0.94 and faithfulness of 0.85. These results show that KITE’s multi-stage retrieval pipeline surfaced course-specific material relevant to the questions and that its responses remained closely grounded in that retrieved context. At the same time, the lower factual correctness score (0.45) warrants careful interpretation, particularly relative to answer similarity (0.76). Prior work has noted that RAGAs-style claim matching and score stability can vary across response formulations and evaluation settings~\cite{roychowdhury2024evaluation, springerragas2025}. In light of prior work, the 0.31-point gap observed in our results may reflect limitations of factual correctness as a reference-based metric for evaluating KITE’s pedagogically framed responses, although our evaluation does not isolate the source of that discrepancy.
\\

\textbf{Pedagogical effectiveness and design implications.} Following the RAGAs evaluation, the simulated student and expert evaluations examined whether KITE’s feedback supported stronger revised answers, and was judged to be pedagogically appropriate. Specifically, the 88.89\% improvement rate in the simulated student pipeline, together with the strong expert rubric scores for scaffolding, guidance, coherence, and tone, support that KITE’s feedback can support stronger follow-up answers on procedural and tracing questions. This emphasis on feedback quality is consistent with prior survey work on LLM applications in programming education, which argues that the educational value of these systems depends on aligning model capabilities with pedagogical goals, including the use of scaffolding and feedback strategies~\cite{pitts2025survey}. In KITE, this alignment is reflected in pairing retrieval-grounded generation with feedback strategies designed for different forms of student support, such as direct explanations, or algorithmic-tracing guidance. While these findings are encouraging, the current evaluation design limits their generalizability, as discussed in Section \ref{limitations}.

\subsection{Limitations and Future Work}
\label{limitations}

This study has limitations that motivate future work. First, our evaluation of factual correctness is constrained by limitations of the RAGAs framework. KITE produces pedagogically framed explanations that may paraphrase or elaborate on course material, whereas RAGAs decomposes each response into atomic claims and uses NLI-style entailment to assess agreement with a single instructor-authored answer. With this, responses that are semantically aligned with the expected answer but differ in phrasing, detail, or framing may receive lower factual correctness scores. The 0.31-point gap between factual correctness (0.45) and answer similarity (0.76) in Table~\ref{tab:ragas} is consistent with this concern, although our evaluation does not isolate the source of that discrepancy. Answer similarity, which is based on semantic embeddings, remains substantially higher and shows low variance ($\sigma = 0.09$), indicating relatively stable semantic alignment across questions. We therefore treat answer similarity as the primary quality indicator and note that using multiple human-written answers could reduce this limitation in future evaluations. Related concerns have been noted in prior work: Roychowdhury et al.~\cite{roychowdhury2024evaluation} discuss limitations in how RAGAs decomposes and assigns statements during metric computation, while Antal and Buza~\cite{springerragas2025} show that RAGAs-based evaluation outcomes vary across question types and retrieval conditions. Future evaluations could use richer answer sets and metrics.

Second, the simulated student pipeline relies on a single LLM, Meta-Llama-3.1-70B-Instruct, as a proxy for student behavior. As a result, improvement between Round 1 and Round 2 should be interpreted as evidence that KITE's feedback makes a stronger answer more recoverable, not as direct evidence of genuine student learning. Real learners may differ substantially in both the magnitude and pattern of improvement. Although this design is useful for early-stage evaluation, it cannot substitute for classroom evidence. A necessary next step is deployment with real students, including pre- and post-interaction assessments, analysis of revision behavior over time, and closer examination of how learners engage with feedback.

Third, the expert evaluation covers a relatively limited set of interaction cases, which constrains the generalizability of the findings. Although inter-rater agreement was strong ($\kappa = 0.88$), judgments of answer improvement and pedagogical quality still involve subjectivity, and the sample size limits precision. Future work should expand the annotation set and use finer-grained scoring schemes to better capture variation in feedback quality and answer improvement across question and error types.

\section{Conclusion} 

We presented KITE, a RAG-based intelligent tutoring system that combines a five-stage retrieval pipeline with intent-aware pedagogical response generation. KITE adapts its responses to the type of student query, providing grounded explanations for factual questions and Socratic scaffolding for procedural and reasoning tasks. To evaluate these response modes, we introduced a two-part evaluation framework. RAGAs metrics assess retrieval quality, while a simulated student pipeline examines whether KITE’s feedback supports improved responses on procedural and tracing questions. Expert review using a structured rubric further evaluates the pedagogical quality of KITE’s feedback and verifies improvement in students’ revised answers. This work contributes an intent-aware tutoring architecture and an evaluation approach for RAG systems with mixed response strategies.

\section*{Acknowledgements}
This research was supported by the U.S. National Science Foundation (NSF) under Grant \#2426837. Any opinions, findings, and conclusions expressed in this material are those of the authors and do not necessarily reflect views of the NSF. This work was additionally supported by Research Council of Finland grants \#356114 and \#367787. 

\bibliography{custom}


\end{document}